\documentclass[11pt,letterpaper]{article}
\usepackage[hyperref]{acl2017}
\usepackage{times}
\usepackage{latexsym}
\usepackage{amsmath}
\usepackage{amssymb}
\usepackage{graphicx}
\usepackage{ifthen}
\usepackage{color}
\usepackage[utf8]{inputenc}
\usepackage{url}

\usepackage{xspace}
\usepackage{booktabs}
\usepackage{arydshln}
\usepackage{enumitem}
\usepackage{listings}
\usepackage{csquotes}
\usepackage{subcaption}
\usepackage[noline,boxed]{algorithm2e}
\SetAlCapSkip{1em}

\usepackage[T1]{fontenc}

\usepackage{oubraces}
\usepackage{tikz}

\newcommand\br{\ensuremath{\mathbf{r}}}

\newcommand\R{\ensuremath{\mathbb{R}}} 
\newcommand\eqdef{\ensuremath{\stackrel{\rm def}{=}}} 

\newcommand\refsec[1]{Section~\ref{sec:#1}}

\newcommand\reffig[1]{Figure~\ref{fig:#1}}

\newcommand\reftab[1]{Table~\ref{tab:#1}}

\newcommand\refalg[1]{Algorithm~\ref{alg:#1}}

\newcommand\Section[2]{\section{#2}\label{sec:#1}}
\newcommand\Subsection[2]{\subsection{#2}\label{sec:#1}}

\ifthenelse{\isundefined{\definition}}{\newtheorem{definition}{Definition}}{}
\ifthenelse{\isundefined{\assumption}}{\newtheorem{assumption}{Assumption}}{}
\ifthenelse{\isundefined{\hypothesis}}{\newtheorem{hypothesis}{Hypothesis}}{}
\ifthenelse{\isundefined{\proposition}}{\newtheorem{proposition}{Proposition}}{}
\ifthenelse{\isundefined{\theorem}}{\newtheorem{theorem}{Theorem}}{}
\ifthenelse{\isundefined{\lemma}}{\newtheorem{lemma}{Lemma}}{}
\ifthenelse{\isundefined{\corollary}}{\newtheorem{corollary}{Corollary}}{}
\ifthenelse{\isundefined{\alg}}{\newtheorem{alg}{Algorithm}}{}
\ifthenelse{\isundefined{\example}}{\newtheorem{example}{Example}}{}

\newif\ifshowcomments
\showcommentstrue
\ifshowcomments
\newcommand\pl[1]{\textcolor{red}{[PL: #1]}}
\definecolor{CMpurple}{rgb}{0.6,0.18,0.64}

\newcommand\sidaw[1]{\textcolor{blue}{[sidaw: #1]}}
\newcommand\TODO[1]{\textcolor{red}{[TODO: #1]}}
\else
\newcommand\pl[1]{}
\newcommand\sidaw[1]{}
\newcommand\TODO[1]{}
\fi

\newcommand{\autocap}[1]{
  \ifnum\spacefactor=3000
    \expandafter\MakeUppercase\fi%
  #1}

\newcommand{\action}{\ensuremath{A}} 
\newcommand{\rel}{\ensuremath{R}\xspace} 
\newcommand{\valueset}{\ensuremath{S}\xspace} 
\newcommand{\itemset}{\ensuremath{S}\xspace} 
\newcommand{\colors}{\ensuremath{C}\xspace} 
\providecommand{\gram}{{\ensuremath{\rightarrow} }}

\newcommand{\score}{\operatorname{score}}
\newcommand{\chart}{\operatorname{chart}}
\newcommand\exec[2]{\ensuremath{\llbracket#2\rrbracket_{#1}}}

\providecommand{\T}{\mathsf{T}} 

\providecommand{\utt}[1]{{`\fontfamily{cmss}\selectfont#1}'}
\providecommand{\core}[1]{`{\fontfamily{cmss}\selectfont#1}'}
\providecommand{\stp}[1]{{\fontfamily{cmss}\selectfont #1} -- }
\providecommand{\uttnq}[1]{{\fontfamily{cmss}\selectfont #1}}

\providecommand{\Pd}[1]{\operatorname{#1}}

\usepackage{stmaryrd}

\newcommand{\reg}[1]{||#1||_1}

\newcommand{\forcecommand}[3][0]{
  \providecommand{#2}{}
  \renewcommand{#2}[#1]{#3}
}
\forcecommand {\maximize} {\operatorname{maximize}}
\forcecommand {\minimize} {\operatorname{minimize}}

\DeclareMathOperator*{\argmax}{argmax}

\newcommand*{\argmaxl}{\argmax\limits}

\forcecommand {\subjecto} {\operatorname{subject\ to}}
\forcecommand{\lone} {\ensuremath{\ell_1}}
\forcecommand{\ltwo} {\ensuremath{\ell_2}}

\aclfinalcopy

\title{Naturalizing a Programming Language via Interactive Learning}

\author{
  Sida I. Wang, Samuel Ginn, Percy Liang, Christopher D. Manning\\
  Computer Science Department \\
  Stanford University \\
  {\tt \{sidaw, samginn, pliang, manning\}@cs.stanford.edu}
}

\begin{document}

\maketitle

\begin{abstract}
Our goal is to create a convenient natural language interface for
performing well-specified but complex actions
such as analyzing data, manipulating text, and querying databases.
However, existing natural language interfaces for such tasks are quite primitive
compared to the power one wields with a programming language.
To bridge this gap, we start with a
core programming language and allow users to
``naturalize'' the core language incrementally
by defining alternative, more natural syntax and
increasingly complex concepts in terms of
compositions of simpler ones.
In a voxel world, we show that a community of users
can simultaneously
teach a common system a diverse language
and use it to build hundreds of complex voxel structures.
Over the course of three days,
these users went from using only the core language
to using the naturalized language in 85.9\% of
the last 10K utterances.
%

\end{abstract}


\begin{figure}[ht!]

		\begin{subfigure}[b]{0.55\columnwidth}
				\begin{center}
					\includegraphics[width=\textwidth]{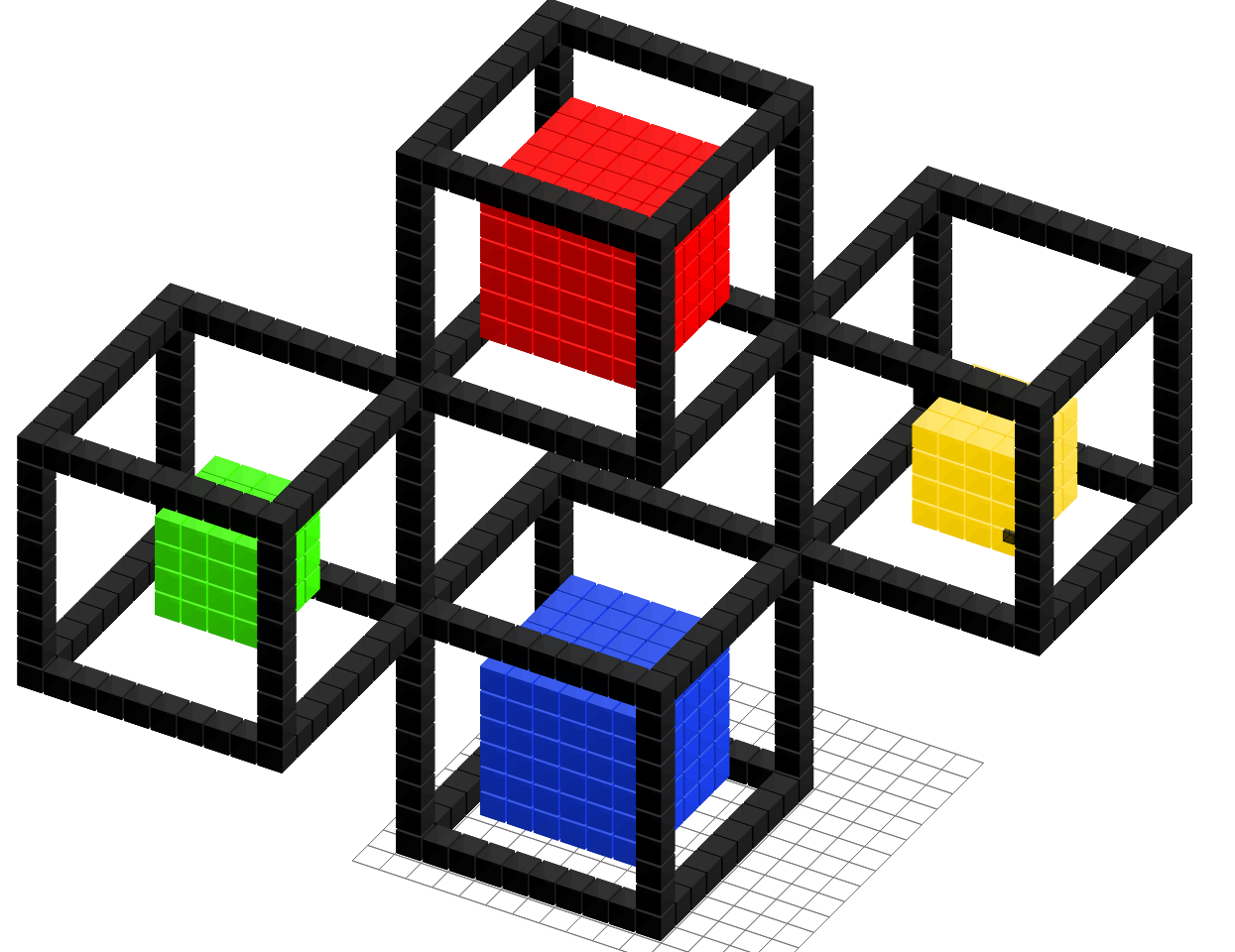}
			\end{center}
		\end{subfigure}\hfill
		\begin{subfigure}{0.45\columnwidth}
			\vspace{-7em}
		\small{
		\textbf{Cubes: }
		\stp{initial}
		\stp{select left 6}
		\stp{select front 8}
		\stp{black 10x10x10 frame}
		\stp{black 10x10x10 frame}
		\stp{move front 10}
		\stp{red cube size 6}
		\stp{move bot 2}
		\stp{blue cube size 6}
		\stp{green cube size 4}
		(some steps are omitted)
		}
		\end{subfigure}

	\begin{subfigure}[b]{0.52\columnwidth}
		\includegraphics[width=\textwidth]{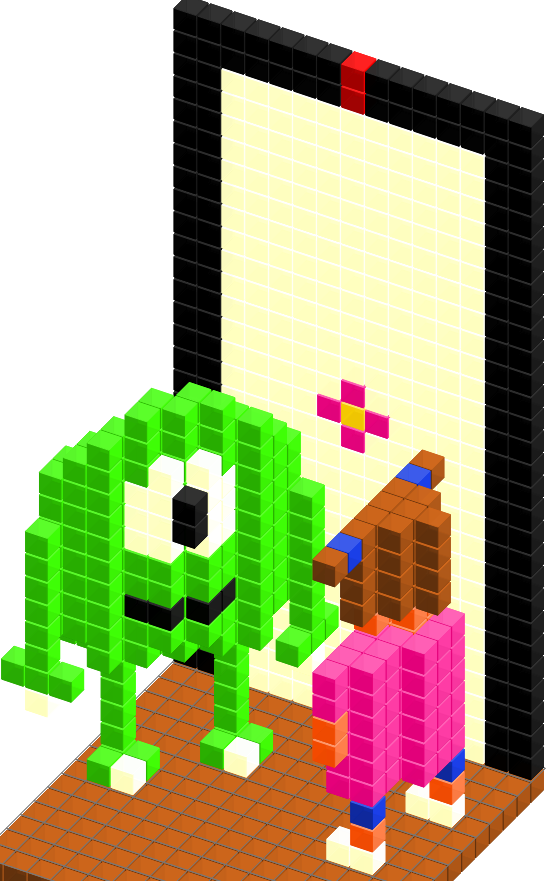}
	\end{subfigure}
	\hfill
  \begin{subfigure}[b]{0.46\columnwidth}
    \includegraphics[width=\textwidth]{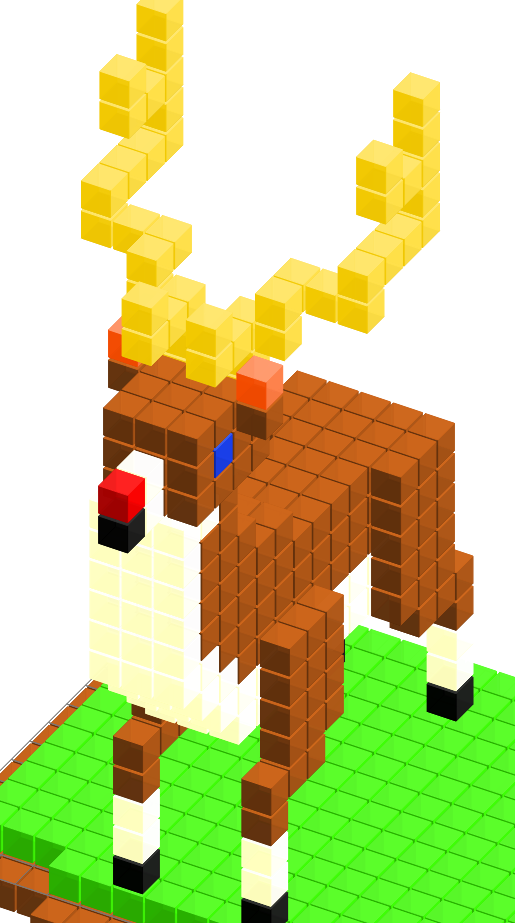}
  \end{subfigure}

	\begin{subfigure}{1\columnwidth}
		\vspace{0.3em}
		\raggedright
		\small{
		\textbf{Monsters, Inc: }
		\stp{initial}
		\stp{move forward}
		\stp{add green monster}
		\stp{go down 8}
		\stp{go right and front}
		\stp{add brown floor}
		\stp{add girl}
		\stp{go back and down}
		\stp{add door}
		\stp{add black column 30}
		\stp{go up 9}
		\stp{finish door}
		(some steps for moving are omitted)
		}
  \end{subfigure}

	\begin{subfigure}{1\columnwidth}
	\vspace{0.3em}
	\small{
	\textbf{Deer: }
	\stp{initial}
	\stp{bird's eye view}
	\stp{deer head; up; left 2; back 2; \{ left antler \}; right 2; \{right antler\}}
	\stp{down 4; front 2; left 3; deer body; down 6; \{deer leg front\}; back 7; \{deer leg back\}; left 4; \{deer leg back\}; front 7; \{deer leg front\}}
    (some steps omitted) 
	}
  \end{subfigure}
	\vspace{-0.2em}
  \caption{Some examples of users building structures using a naturalized language in Voxelurn:\linebreak
  \textbf{\url{http://www.voxelurn.com}}
  }
	\label{fig:structs}
\end{figure}
%

\section{Introduction}


In tasks such as
analyzing and plotting data \cite{gulwani2014nlyze}, 
querying databases \cite{zelle96geoquery,berant2013freebase},
manipulating text \cite{kushman2013regex},
or controlling the Internet of Things \cite{campagna2017almond} and robots \cite{tellex2011understanding},
people need computers to perform well-specified
but complex actions.
To accomplish this,
one route is to use a programming language,
but this is inaccessible to most and
can be tedious even for experts because
the syntax is uncompromising and all statements have to be precise.
Another route is to convert natural
language into a formal language,
which has been the subject of work in semantic parsing
\citep{zettlemoyer05ccg,artzi11conversations,artzi2013weakly,pasupat2015compositional}. 
However, the capability of semantic parsers
is still quite primitive
compared to the power one wields with a programming language.
This gap is increasingly limiting the potential of both text
and voice interfaces as they become more ubiquitous and desirable.

In this paper, we propose bridging this gap with
an interactive language learning process which we call \emph{naturalization}.
Before any learning, we seed a system with a core programming language
that is always available to the user.
As users instruct the system to perform actions,
they augment the language by \emph{defining} new utterances
---
e.g., the user can explicitly tell the computer that
\utt{X} means \utt{Y}.
Through this process, users gradually and interactively teach
the system to understand the language that they
\emph{want to use}, rather than the core language
that they are forced to use initially.
While the first users have to learn the core
language,
later users can make use of everything that is already taught.
This process accommodates both
users' preferences and
the computer action space,
where the final language is both
interpretable by the computer and easier to produce by human users.

Compared to interactive language learning with
weak denotational supervision \cite{wang2016games},
definitions are
critical for learning complex actions
(\reffig{structs}).
Definitions equate a novel utterance to a sequence of utterances
that the system already understands.
For example, \utt{go left 6 and go front}
might be defined as \utt{repeat 6 [go left]; go front},
which eventually can be traced back to the expression
\utt{repeat 6 [select left of this]; select front of this} in the core language.
Unlike function definitions in programming languages,
the user writes concrete values rather than explicitly declaring arguments.
The system automatically extracts arguments and learns to produce the correct generalizations.
For this, we propose a grammar induction algorithm tailored to
the learning from definitions setting.
Compared to standard machine learning, say from demonstrations,
definitions provide a much more powerful learning signal:
the system is told directly
that \utt{a 3 by 4 red square} is \utt{3 red columns of height 4},
and does not have to infer how to generalize
from observing many structures of different sizes.

We implemented a system called Voxelurn,
which is a command language interface for a voxel world
initially equipped with a programming language
supporting conditionals, loops, and variable scoping etc.
We recruited 70 users from Amazon Mechanical Turk
to build 230 voxel structures using our system.
All users teach the system at once,
and what is learned from one user can be used by another user.
Thus a \emph{community} of users evolves the language
to becomes more efficient over time, in a distributed way, through interaction.
We show that the user community defined many new utterances%
---short forms,
alternative syntax,
and also complex concepts such as \utt{add green monster, add yellow plate 3 x 3}.
As the system learns,
users increasingly prefer to use the naturalized language over
the core language: 85.9\% of
the last 10K accepted utterances
are in the naturalized language.

\begin{figure}[h!]
	\begin{subfigure}[b]{\columnwidth}
		\includegraphics[width=\textwidth]{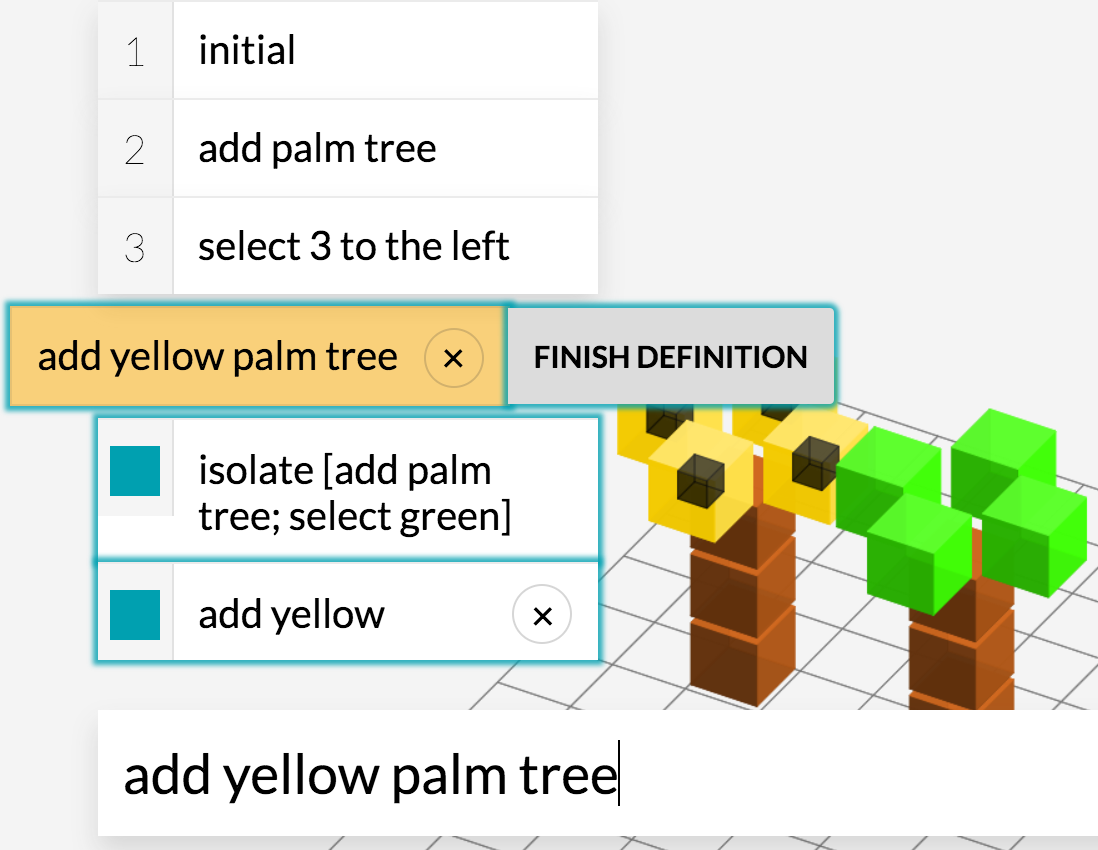}
	\end{subfigure}
  \caption{Interface used by users to enter utterances and create definitions.}
	\label{fig:interface}
\end{figure}

\newcommand\exsep{\,\,\uttnq{|}\,\,}

\begin{table*}
  \begin{center}
\begin{tabular}{rll}
\textbf{Rule(s)} & \textbf{Example(s)} & \textbf{Description} \\
\midrule

\action \gram \action; \action & \uttnq{select left; add red}
&   perform actions sequentially \\

\action \gram repeat $N$ \action & \uttnq{repeat 3-1 add red top}
 &  repeat action $N$ times\\

\action \gram if \itemset  \action & \uttnq{if has color red [select origin]}
 &  action if \valueset is non-empty\\

 \action \gram while \itemset  \action & \uttnq{while not has color red [select left of this]}
  &  action while \valueset is non-empty\\

\action \gram foreach \itemset \action  & \uttnq{foreach this [remove has row row of this]}
 &  action for each item in $S$ \\

\action \gram [\action] & \uttnq{[select left or right; add red; add red top] }
  & group actions for precedence \\

\action \gram \string{\action\string} & \uttnq{\{select left; add red\} }
 & scope only selection \\

 \action \gram isolate \action & \uttnq{isolate [add red top; select has color red]}
  & scope voxels and selection \\


\midrule
\action \gram select \itemset & \uttnq{select all and not origin}
 & set the selection \\
 \action \gram remove \itemset & \uttnq{remove has color red}
 & remove voxels \\
\action \gram update \rel \valueset & \uttnq{update color [color of left of this]}
 & change property of selection \\
\itemset & \uttnq{this} & current selection \\
\itemset & \uttnq{all} \exsep \uttnq{none} \exsep \uttnq{origin} & all voxels, empty set, $(0,0)$ \\

\midrule
\rel of \itemset \exsep
has \rel \itemset
& \uttnq{has color red or yellow} \exsep \uttnq{has row [col of this]}
& lambda DCS joins \\

not \itemset \exsep
\itemset and \itemset \exsep
\itemset or \itemset
& \uttnq{this or left and not has color red}
& set operations \\

$N$ \exsep $N$+$N$ \exsep $N$-$N$
& \uttnq{1,\ldots,10} \exsep \uttnq{1+2} \exsep \uttnq{row of this + 1}
& numbers and arithmetic \\

argmax \rel \itemset \exsep
argmin \rel \itemset
& \uttnq{argmax col has color red}
& superlatives \\

\midrule
\rel
& \uttnq{color} \exsep  \uttnq{row} \exsep  \uttnq{col} \exsep \uttnq{height} \exsep \uttnq{top} \exsep \uttnq{left} \exsep $\cdots$ & voxel relations \\

\colors
& \uttnq{red} \exsep  \uttnq{orange} \exsep  \uttnq{green} \exsep \uttnq{blue} \exsep \uttnq{black} \exsep $\cdots$ &  color values \\

$D$
& \uttnq{top} \exsep  \uttnq{bot} \exsep  \uttnq{front} \exsep \uttnq{back} \exsep \uttnq{left} \exsep \uttnq{right} & direction values  \\

\itemset \gram very $D$ of \itemset
& \uttnq{very top of very bot of has color green} & syntax sugar for argmax \\

\action \gram
add \colors [$D$] \exsep
move $D$
& \uttnq{add red} \exsep  \uttnq{add yellow bot} \exsep  \uttnq{move left}
& add voxel, move selection


\end{tabular}
\caption{Grammar of the core language (DAL),
which includes actions ($A$), relations ($R$), and sets of values ($S$).
The grammar rules are grouped into four categories.
From top to bottom:
domain-general action compositions,
actions using sets,
lambda DCS expressions for sets,
and domain-specific relations and actions.
}
\label{tab:core}
\end{center}
\vspace{-1.7em}
\end{table*}

\section{Voxelurn}
\label{sec:core}

\paragraph{World.}

A world state in Voxelurn contains a set of voxels,
where each voxel has relations \core{row}, \core{col}, \core{height}, and \core{color}.
There are two domain-specific actions, \core{add} and \core{move},
one domain-specific relation \core{direction}.
In addition, the state contains a selection,
which is a set of positions.
While our focus is Voxelurn,
we can think more generally about the world as a set of objects equiped with relations
--- events on a calendar, cells of a spreadsheet, or lines of text.
%

\paragraph{Core language.}
The system is born understanding a core language called Dependency-based Action Language (DAL),
which we created (see \reftab{core} for an overview).

The language composes actions
using the usual but expressive
control primitives such as
\core{if}, \core{foreach}, \core{repeat}, etc.
Actions usually take sets as arguments,
which are represented using
lambda dependency-based compositional semantics (lambda DCS)
expressions \cite{liang2013lambdadcs}.
Besides standard set operations like union, intersection and complement,
lambda DCS leverages the tree dependency structure common in natural language:
for the relation \core{color}, \core{has color red} refers to the
set of voxels that have color red, and its reverse \core{color of has row 1} refers
to the set of colors of voxels having row number 1.
Tree-structured joins can be chained without using any variables, e.g.,
\core{has color [yellow or color of has row 1]}.

We protect the core language from being redefined
so it is always precise and usable.%
\footnote{Not doing so
resulted in ambiguities that propagated uncontrollably, e.g., once \utt{red}
can mean many different colors.}
In addition to expressivity, the core language \emph{interpolates} well with natural language.
We avoid explicit variables by using a \emph{selection},
which serves as the default argument for most actions.%
\footnote{The selection
is like the turtle in LOGO, but can be a set.}
For example,
\core{select has color red; add yellow top; remove} adds yellow on top of
red voxels and then removes the red voxels.

To enable the building of more complex structures in a more modular way,
we introduce a notion of \emph{scoping}.
Suppose one is operating on one of the palm trees in \reffig{interface}.
The user might want to use \core{select all} to select only the voxels in that tree
rather than all of the voxels in the scene.
In general, an action $A$ can be viewed as taking a set of voxels $v$ and a selection $s$,
and producing an updated set of voxels $v'$ and a modified selection $s'$.
The default scoping is \core{[$A$]}, which is the same as \core{$A$} and returns $(v',s')$.
There are two constructs that alter the flow:
First, \core{\{$A$\}} takes $(v,s)$ and returns $(v',s)$,
thus restoring the selection.
This allows $A$ to use the selection as a temporary variable without affecting the rest of the program.
Second, \core{isolate [$A$]} takes $(v,s)$, calls $A$ with $(s,s)$ (restricting
the set of voxels to just the selection) and returns $(v'',s)$,
where $v''$ consists of voxels in $v'$ and voxels in $v$
that occupy empty locations in $v'$.
This allows $A$ to focus only on the selection (e.g., one of the palm trees).
Although scoping can be explicitly controlled via \core{[\,]}, \core{isolate}, and \core{\{\,\}},
it is an unnatural concept for non-programmers.
Therefore when the choice is not explicit,
the parser generates all three possible scoping interpretations,
and the model learns which is intended based on the user, the rule, and potentially the context.




\section{Learning interactively from definitions}

The goal of the user is to build a structure in Voxelurn.
In \citet{wang2016games},
the user provided interactive supervision to the system by selecting from a list of candidates.
This is practical when there are less than tens of candidates,
but is completely infeasible for a complex action space such as Voxelurn.
Roughly, 10 possible colors over
the $3\times3\times4$ box containing the palm tree in
\reffig{interface} yields $10^{36}$ distinct denotations,
and many more programs.
Obtaining the structures in \reffig{structs} by selecting candidates alone would be infeasible.

This work thus uses \emph{definitions}
in addition to selecting candidates as the supervision signal.
Each definition consists of a \emph{head} utterance
and a \emph{body}, which is a sequence of utterances that the system understands.
One use of definitions is paraphrasing and defining alternative syntax,
which helps naturalize the core language
(e.g., defining \utt{add brown top 3 times} as \core{repeat 3 add brown top}).
The second use is building up complex concepts hierarchically.
In \reffig{interface}, \utt{add yellow palm tree} is defined as a sequence of steps for building the palm tree.
Once the system understands an utterance,
it can be used in the body of other definitions.
For example, \reffig{palmtree_example} shows the full definition tree of \core{add palm tree}.
Unlike function definitions in a programming language,
our definitions do not specify the exact arguments;
the system has to learn to extract arguments to achieve the correct generalization.



\setlength{\algomargin}{0.5em}
\begin{algorithm}
 \SetKwBlock{Begin}{def:}{defined}
 \SetAlgoNoEnd
\SetAlgorithmName{definition}{definition}\quad
\Begin(\uttnq{add palm tree}) {
	\Begin(\uttnq{brown trunk height 3}) {
		\Begin(\uttnq{add brown top 3 times}){
		\uttnq{repeat 3 [add brown top]}
		}
	}
	\Begin(\uttnq{go to top of tree}) {
	\uttnq{select very top of has color brown}
	}
	\Begin(\uttnq{add leaves here}) {
		\Begin(\uttnq{select all sides}){
			\uttnq{select left or right or front or back}
		}
		\uttnq{add green}
	}
}
\captionof{figure}
{Defining \utt{add palm tree},
tracing back to the core language (utterances without \textbf{def:}).} \label{fig:palmtree_example}
\end{algorithm}


\begin{algorithm}
\SetAlgoNoEnd
\SetAlgorithmName{}{}\quad
 \Begin(execute $x$:){
	\lIf{$x$ does not parse}
	{define $x$}
	\lIf{user rejects all parses}{define $x$}
	execute user choice
 }\vspace{0.5em}
 \Begin(define $x$:){
 \Repeat ( starting with $X \gets [\,]$){user accepts $X$ as the def'n of $x$}{
  user enters $x'$\;
	\lIf{$x'$ does not parse}{define $x'$}
	\lIf{user rejects all $x'$}{define $x'$}
   $X \gets [X; x']$\;
 }
 }
\captionof{figure}{When the user enters an utterance,
the system tries to parse and execute it,
or requests that the user define it.
} \label{fig:def_process}
\end{algorithm}

The interactive definition process is described in \reffig{def_process}.
When the user types an utterance $x$, the system parses $x$ into a list of candidate programs.
If the user selects one of them (based on its denotation), then the system executes the resulting program.
If the utterance is unparsable or the user rejects all candidate programs,
the user is asked to provide the definition body for $x$.
Any utterances in the body not yet understood can be defined recursively.
Alternatively, the user can first execute
a sequence of commands $X$,
and then provide a head utterance for body $X$.


When constructing the definition body,
users can type utterances with multiple parses;
e.g., \utt{move forward} could either modify the selection (\core{select front}) or move the voxel (\core{move front}).
Rather than propagating this ambiguity to the head,
we force the user to commit to one interpretation by selecting a particular candidate.
Note that we are using interactivity to control the exploding ambiguity.

\section{Model and learning}
\label{sec:methods}

Let us turn to how the system learns and predicts.
This section contains prerequisites before we
describe definitions and grammar induction in \refsec{grammar}.

\paragraph{Semantic parsing.}

Our system is based on a semantic parser that maps utterances $x$ to programs $z$,
which can be executed on the current state $s$ (set of voxels and selection)
to produce the next state $s' = \exec{s}{z}$.
Our system is implemented as the interactive package in SEMPRE \citep{berant2013freebase};
see \citet{liang2016executable} for a gentle exposition.

A \emph{derivation} $d$ represents the process by which an utterance $x$ turns into a program $z = \Pd{prog}(d)$.
More precisely, $d$ is a tree where each node contains
the corresponding span of the utterance $(\Pd{start}(d), \Pd{end}(d))$,
the grammar rule $\Pd{rule}(d)$,
the grammar category $\Pd{cat}(d)$,
and a list of child derivations $[d_1, \ldots, d_n]$.

Following \citet{zettlemoyer05ccg},
we define a log-linear model over derivations $d$
given an utterance $x$ produced by the user $u$:
\begin{align}
p_\theta(d \mid x,u) &\propto \exp(\theta^\T \phi(d,x,u)),
\end{align}
where $\phi(d,x,u) \in \R^p$ is a feature vector and $\theta \in \R^p$ is a
parameter vector.  The user $u$ does not appear in previous work on semantic parsing,
but we use it to personalize the semantic parser trained on the community.


We use a standard chart parser to construct a chart.
For each chart cell, indexed by the start and end indices of a span,
we construct a list of partial derivations recursively by selecting
child derivations from subspans and applying a grammar rule.
The resulting derivations are sorted by model score and only the top $K$ are kept.
We use $\Pd{chart}(x)$ to denote the set of all partial derivations across all chart cells.
The set of grammar rules starts with the set of rules for the core language (\reftab{core}),
but grows via grammar induction when users add definitions (\refsec{grammar}).
Rules in the grammar are stored in a trie based on the right-hand side to enable better
scalability to a large number of rules.


\label{sec:features}
\begin{table}
\begin{tabular}{l|l}
 \hline
 Feature & Description  \\
\hline
Rule.ID& ID of the rule \\
Rule.Type & core?, used?, used by others? \\
 \hline
 Social.Author & ID of author \\
 Social.Friends & (ID of author, ID of user) \\
 Social.Self & rule is authored by user? \\
 \hline
 Span & (left/right token(s), category) \\
Scope & type of scoping for each user
\end{tabular}
\caption{Summary of features.}
\label{tab:features}
\end{table}

\paragraph{Features.}

Derivations are scored using a weighted combination of features.
There are three types of features,
summarized in \reftab{features}.

\emph{Rule features} fire on each rule used to construct a derivation.
ID features fire on specific rules (by ID).
Type features track whether a rule is part of the core language or induced,
whether it has been used again after it was defined,
if it was used by someone other than its author,
and if the user and the author are the same
($5 +\text{\#rules}$ features).

\emph{Social features} fire on properties of rules that capture
the unique linguistic styles of different users and their interaction with each other.
Author features capture the fact that some users provide better, and more
generalizable definitions that tend to be accepted.
Friends features are cross products of author ID and user ID,
which captures whether rules from a particular author
are systematically preferred or not by the current
user, due to stylistic similarities or differences
($\text{\#users}+\text{\#users}\times\text{\#users}$ features).


\emph{Span features}
include conjunctions of the category of the derivation and
the leftmost/rightmost token on the border of the span.
In addition, span features include conjunctions of the category of the derivation and
the 1 or 2 adjacent tokens just outside of the left/right border of the span.
These capture a weak form of
context-dependence that is generally helpful
($< \approx V^4\times\ \text{\#cats}$ features for a vocabulary of size $V$).

\emph{Scoping features} track how the community,
as well as individual users,
prefer each of the 3 scoping choices (none, selection only \utt{\{A\}}, and voxels+selection \utt{isolate \{A\}}),
as described in \refsec{core}.
3 global indicators, and 3 indicators for each user fire every time
a particular scoping choice is made
($3+3\times\text{\#users}$ features).

\paragraph{Parameter estimation.}
When the user types an utterance, the system generates a list of candidate next states.
When the user chooses a particular next state $s'$ from this list,
the system performs an online AdaGrad update \citep{duchi10adagrad}
on the parameters $\theta$ according to the gradient of the following loss function:
\begin{align*}
  \label{eqn:loss}
  -\log \sum_{d : \exec{s}{\Pd{prog}(d)} = s'} p_\theta(d \mid x, u) + \lambda \reg{\theta}, \\
\end{align*}
which attempts to increase the model probability on derivations whose programs
produce the next state $s'$.

\Section{grammar}{Grammar induction}

Recall that the main form of supervision is via user definitions,
which allows creation of user-defined concepts.
In this section, we show how to turn these definitions into new grammar rules
that can be used by the system to parse new utterances.

Previous systems of grammar induction for semantic parsing
were given utterance-program pairs $(x,z)$.
Both the GENLEX \citep{zettlemoyer05ccg} and higher-order unification \citep{kwiatkowski10ccg} algorithms
over-generate rules that liberally associate parts of $x$ with parts of $z$.
Though some rules are immediately pruned, many spurious rules are undoubtedly
still kept.
In the interactive setting, we must keep the number of candidates small to avoid a bad user experience,
which means a higher precision bar for new rules.

Fortunately, the structure of definitions makes the grammar induction task easier.
Rather than being given an utterance-program $(x,z)$ pair,
we are given a definition, which consists of an utterance $x$ (head)
along with the body $X = [x_1, \ldots, x_n]$, which is a sequence of utterances.
The body $X$ is fully parsed into a derivation $d$,
while the head $x$ is likely only partially parsed.
These partial derivations are denoted by $\Pd{chart}(x)$.



At a high-level, we find \emph{matches}---partial derivations $\Pd{chart}(x)$ of the head $x$
that also occur in the full derivation $d$ of the body $X$.
A grammar rule is produced by substituting any set of non-overlapping
matches by their categories.
As an example, suppose the user defines
\begin{align*}
\text{\utt{add red top times 3} as \core{repeat 3 [add red top]}}.
\end{align*}
Then we would be able to induce the following two grammar rules:
\begin{align*}
A\ \gram\ & \text{add $C$ $D$ times $N$ }: \\
	& \lambda C D N. \text{\uttnq{repeat $N$ [add $C$ $D$]}}\\
A\ \gram\ & \text{$A$ times $N$}: \\
	& \lambda A N. \text{\uttnq{repeat $N$ [$A$]}}
\end{align*}
The first rule substitutes primitive values (\utt{red}, \utt{top}, and \utt{3})
with their respective pre-terminal categories ($C$, $D$, $N$).
The second rule contains compositional categories like actions ($A$),
which require some care.
One might expect that greedily substituting the largest matches or the match that covers the largest portion
of the body would work,
but the following example shows that this is not the case:
\begin{align*}
\overunderbraces{&\br{1}{\action_1} & & & \br{1}{\action_1} & & & \br{1}{\action_1}} 
{ & \text{\uttnq{add}} \ \text{\uttnq{red}}\ & \text{\uttnq{left}}  & \  \text{\uttnq{and}}\ \text{\uttnq{here}} = & \text{\uttnq{add}}\ \text{\uttnq{red}}\ & \text{\uttnq{left;}}&  \  & \text{\uttnq{add}}\ \text{\uttnq{red}}} 
{ & \br{2}{\action_2} & & \br{2}{\action_2}} 
\end{align*}
Here, both the highest coverage substitution
($A_1$: \utt{add red}, which covers 4 tokens of the body),
and the largest substitution available ($A_2$: \utt{add red left}) would generalize incorrectly.
The correct grammar rule only substitutes the primitive values (\utt{red}, \utt{left}).

\subsection{Highest scoring abstractions}
\label{sec:packing}


We now propose a grammar induction procedure that optimizes a more global
objective and uses the learned semantic parsing model to choose substitutions.
More formally, let $M$ be the set of partial derivations in the head
whose programs appear in the derivation $d_X$ of the body $X$:
\begin{align*}
  M & \eqdef \{d \in \chart(x) : \\
  & \exists d' \in \Pd{desc}(d_X) \land \Pd{prog}(d) = \Pd{prog}(d') \},
\end{align*}
where $\Pd{desc}(d_X)$ are the descendant derivations of $d_X$.
Our goal is to find a \emph{packing} $P \subseteq M$,
which is a set of derivations corresponding to non-overlapping spans of the head.
We say that a packing $P$ is maximal
if no other derivations may be added to it without creating an overlap.


Let $\Pd{packings}(M)$ denote the set of maximal packings,
we can frame our problem as finding the maximal packing
that has the highest score under our current semantic parsing model:
\begin{align}
P^*_L = \argmaxl_{P \in \Pd{packings}(M);} \sum_{d \in P} \score(d).
\end{align}

Finding the highest scoring packing
can be done using dynamic programming
on $P^*_i$ for $i=0,1,\ldots,L$, where $L$ is the length of $x$ and $P^*_0 = \emptyset$.
Since $d \in M$, $\Pd{start}(d)$ and $\Pd{end}(d)$ (exclusive)
refer to span in the head $x$.
To obtain this dynamic program,
let $D_i$ be the highest scoring maximal packing
containing a derivation ending \emph{exactly} at position $i$ (if it exists):
\begin{align}
D_i & = \{d_i\} \cup P^*_{\Pd{start}(d_i)}, \\
d_i & = \argmax_{d \in M; \Pd{end}(d) = i} \Pd{score}(d \cup P^*_{\Pd{start}(d)} ).
\end{align}
Then the maximal packing of up to $i$ can be defined recursively as
\begin{align}
  P^*_i & = \argmax_{D \in \{D_{s(i)+1}, D_{s(i)+2}, \ldots, D_{i} \}} \Pd{score}(D) \\
s(i) & = \max_{d : \Pd{end}(d) \leq i } \Pd{start}(d),
\end{align}
where $s(i)$ is the largest index such that $D_{s(i)}$ is no longer maximal
for the span $(0, i)$ (i.e. there is a $d \in M$ on the span $\Pd{start}(d
) \geq s(i) \land \Pd{end}(d) \leq i$.

\begin{algorithm}
	\SetAlgoNoEnd
	\SetKwInOut{Input}{Input}\SetKwInOut{Output}{Output}
 \Input{$x,d_X,P^*$}
 \Output{rule}
 $r \gets x$\;
 $f \gets d_X$\;
 \For{ $d \in P^*$}{
  $r \gets r[\Pd{cat}(d) / \Pd{span}(d) ]$
	$f \gets \lambda \Pd{cat}(d). f[ \Pd{cat}(d) / d ]$
 }
  \Return{rule $(\Pd{cat}(d_X) \gram\ r : f)$}

 \caption{Extract a rule $r$ from a derivation $d_X$ of body $X$ and a packing $P^*$.
 Here, $f[t / s]$ means substituting $s$ by $t$ in $f$, with the usual
 care about names of bound variables.}
 \label{alg:getrule}
\end{algorithm}

Once we have a packing $P^*=P^*_L$, we can
go through $d \in P^*$ in order of $\Pd{start}(d)$,
as in \refalg{getrule}.
This generates one high precision rule per
packing per definition. In addition to the highest scoring packing,
we also use a ``simple packing'', which includes only primitive values
(in Voxelurn, these are colors, numbers, and directions).
Unlike the simple packing, the rule induced
from the highest scoring packing does not always generalize correctly.
However, a rule that often generalizes incorrectly should be down-weighted,
along with the score of its packings.
As a result, a different rule might be induced next time,
even with the same definition.
%
%
%

\Subsection{extension-by-alignment}{Extending the chart via alignment}

\refalg{getrule} yields high precision rules,
but fails to generalize in some cases.
Suppose that \utt{move up} is defined as \utt{move top},
where \utt{up} does not parse, and does not match anything.
We would like to infer that \utt{up} means \utt{top}.
To handle this,
we leverage a property of definitions that we have not used thus far: the utterances themselves.
If we align the head and body,
then we would intuitively expect aligned phrases to correspond to the same derivations.
Under this assumption, we can then transplant these derivations from $d_X$
to $\chart(x)$ to create new matches.
This is more constrained than the usual alignment problem (e.g., in machine translation)
since we only need to consider spans of $X$ which corresponds to
derivations in $\Pd{desc}(d_X)$.

\begin{algorithm}
	\SetKwInOut{Input}{Input}\SetKwInOut{Output}{Output}
 \Input{$x, X, d_X$}
  \For{ $d \in \Pd{desc}(d_X)$, $x' \in \Pd{spans}(x)$ } {
     \If{ $\Pd{aligned}(x', d, (x, X))$} {
		 $d' \gets d$\;
		 $\Pd{start}(d') \gets \Pd{start}(x')$\;
		 $\Pd{end}(d') \gets \Pd{end}(x')$\;
     $\Pd{chart}(x) \gets \Pd{chart}(x) \cup d'$
     }
   }
 \caption{Extending the chart by alignment:
  If $d$ is aligned with $x'$ based on the utterance,
 then we pretend that $x'$ should also parse to $d$,
 and $d$ is transplanted to $\Pd{chart}(x)$ as if it parsed from $x'$.}
 \label{alg:ext_by_align}
\end{algorithm}

\refalg{ext_by_align} provides the algorithm for extending the chart via alignments.
The $\Pd{aligned}$ function is implemented using the following two heuristics:
\begin{itemize}
  \item \textbf{exclusion}: if all but 1 pair of short spans (1 or 2 tokens) are matched, the unmatched pair is considered aligned.
\item \textbf{projectivity}: if $d_1, d_2 \in \Pd{desc}(d_X) \cap \chart(x)$, then
$\Pd{ances}(d_1, d_2)$ is aligned to the corresponding span in $x$.
\end{itemize}

With the extended chart,
we can run the algorithm from \refsec{packing} to induce rules.
The transplanted derivations (e.g., \utt{up}) might now form new matches which allows
the grammar induction to induce more generalizable rules.
We only perform this extension when the body consists of one utterance,
which tend to be a paraphrase.
Bodies with multiple utterances tend to be new concepts (e.g., \utt{add green monster}),
for which alignment is impossible.
Because users have to select from candidates parses in the interactive setting,
inducing low precision rules that generate many parses degrade the user experience.
Therefore, we induce alignment-based rules conservatively---only when all but 1 or 2 tokens of the head aligns to the body and vice versa.




\section{Experiments}
\label{sec:experiment}

\begin{figure}
\centering
  \begin{subfigure}[b]{0.25\columnwidth}
    \includegraphics[width=\textwidth]{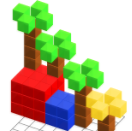}
  \end{subfigure}
\caption{The target used for the qualifier.}
\label{fig:targets}
\end{figure}

\paragraph{Setup.}
Our ultimate goal is to create a community of users who can build interesting structures in Voxelurn
while naturalizing the core language.
We created this community using Amazon Mechanical Turk (AMT) in two stages.
First, we had \emph{qualifier} tasks,
in which an AMT worker was instructed to replicate a fixed target exactly (\reffig{targets}),
ensuring that the initial users
are familiar with at least some of the core language,
which is the starting point of the naturalization process.

Next, we allowed the workers who qualified to enter the second \emph{freebuilding} task,
in which they were asked to build any structure they wanted in 30 minutes.
This process was designed to give users freedom while ensuring quality.
The analogy of this scheme in a real system is that early users (or a small portion of expert users)
have to make some learning investment, so the system can learn and become easier for other users.


\paragraph{Statistics.}
70 workers passed the qualifier task, 
and 42 workers participated in the final free-building experiment.
They built 230 structures.
There were over 103,000 queries consisting of 5,388 distinct token types.
Of these, 64,075 utterances were tried and 36,589 were accepted (so an action was performed).
There were 2,495 definitions combining over 15,000 body utterances
with 6.5 body utterances per head on average (96 max).
From these definitions, 2,817 grammar rules were induced,
compared to less than 100 core rules.
Over all queries, there were
8.73 parses per utterance on average (starting from 1 for core).

\begin{figure}[ht!]
	\vspace{-0.75em}
	\begin{subfigure}[b]{0.9\columnwidth}
		\includegraphics[width=\textwidth]{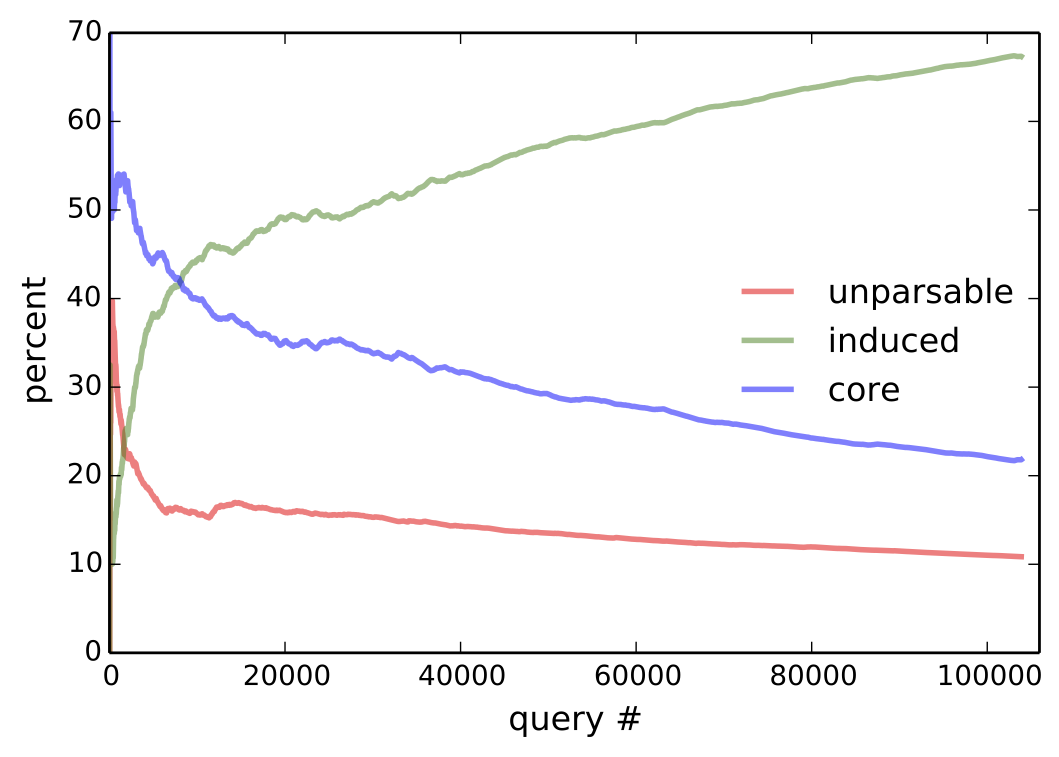} 
	\end{subfigure}
        \vspace{-0.25em}
	\begin{subfigure}[b]{0.9\columnwidth}
		\includegraphics[width=\textwidth]{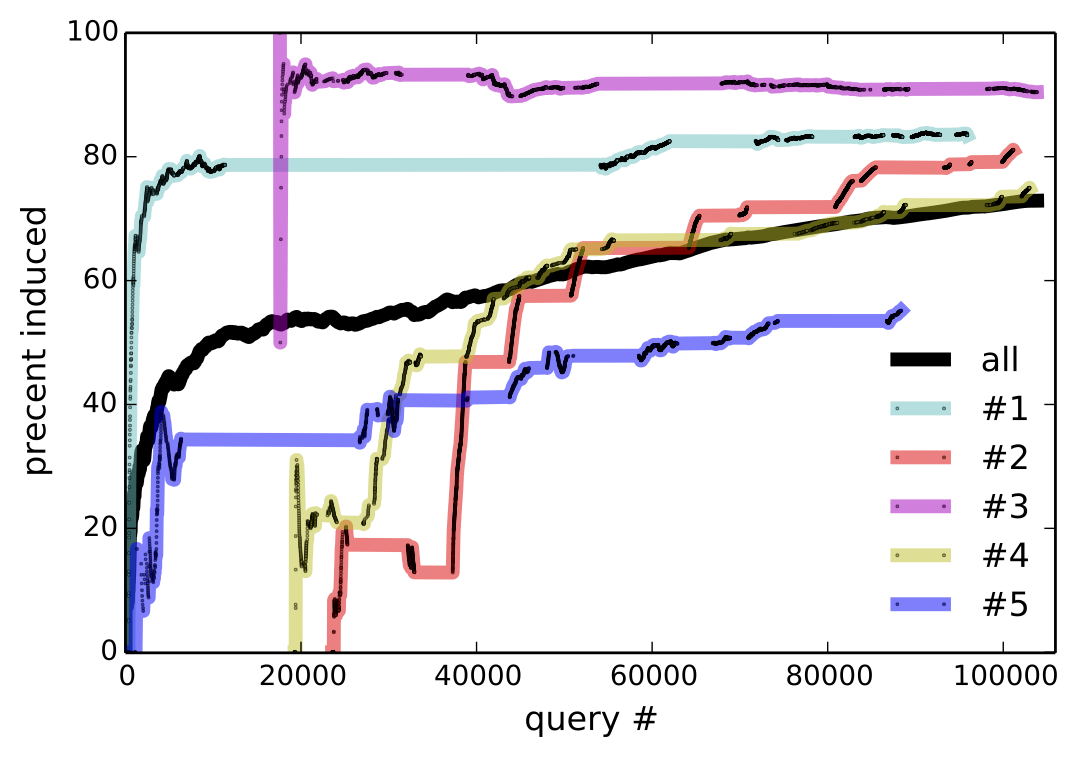}
	\end{subfigure}
	\vspace{-0.5em}
	\begin{subfigure}[b]{0.9\columnwidth}
		\includegraphics[width=\textwidth]{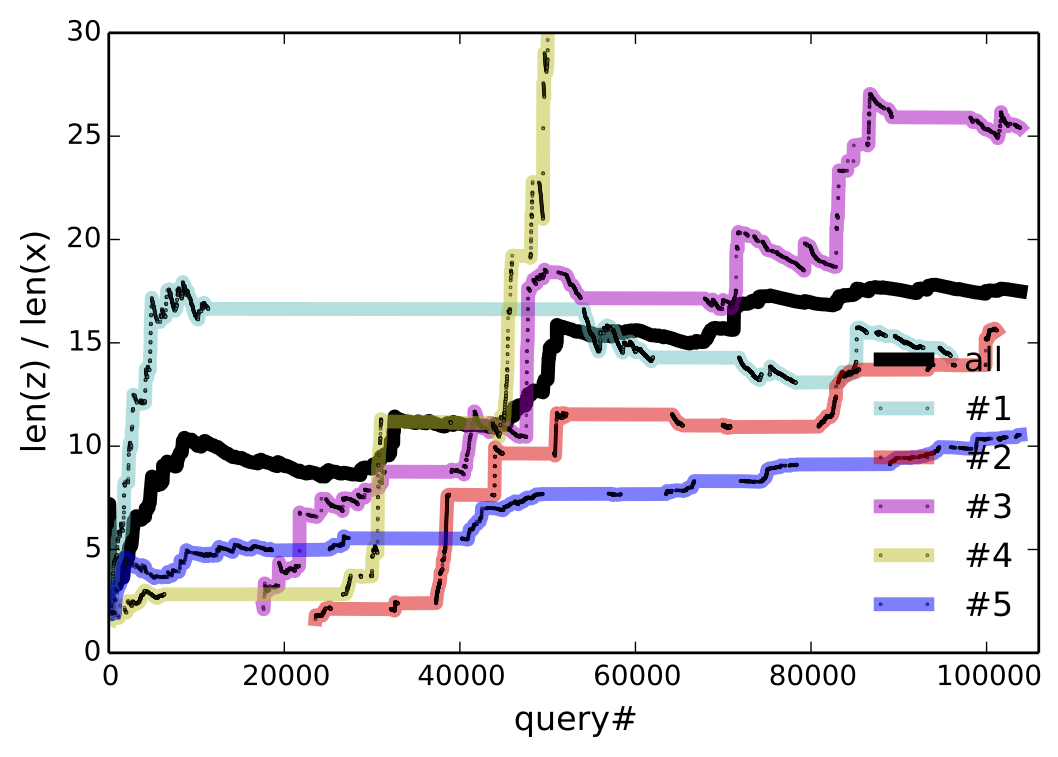}
	\end{subfigure}
	\caption{Learning curves.
	\textbf{Top:} percentage of all utterances that are part of the \emph{core} language, the \emph{induced} language, or \emph{unparsable} by the system.
	\textbf{Middle:} percentage of accepted utterances belonging
        to the induced language, overall and for the 5 heaviest users.
	\textbf{Bottom:} expressiveness measured by the ratio of the length of the program to the length of the corresponding utterance.
	}
	\label{fig:learning_curves}
\end{figure}
\paragraph{Is naturalization happening?} The answer is yes according to \reffig{learning_curves},
which plots the cummulative percentage of utterances that are core,
induced, or unparsable.
To rule out that more induced utterances are getting rejected,
we consider only accepted utterances in the middle of \reffig{learning_curves},
which plots the percentage of induced rules among accepted utterances for the entire community,
as well as for the 5 heaviest users.
Since unparsable utterances cannot be accepted,
accepted core (which is not shown) is the complement of accepted induced.
At the conclusion of the experiment, 72.9\% of all accepted utterances are induced---this
becomes 85.9\% if we only consider the final 10,000 accepted utterances.

\begin{table}[h!]
\begin{tabular}{l}
 \hline
 \textbf{Short forms} \\\hline
 \uttnq{left}, \uttnq{l}, \uttnq{mov left}, \uttnq{go left}, \uttnq{<}, \uttnq{sel left}\\
  \uttnq{br}, \uttnq{black}, \uttnq{blu}, \uttnq{brn}, \uttnq{orangeright}, \uttnq{left3}\\
	\uttnq{add row brn left 5} := \uttnq{add row brown left 5}\\
 \hline
 \textbf{Alternative syntax} \\\hline
 \uttnq{go down and right} := \uttnq{go down; go right}\\
 \uttnq{select orange} := \uttnq{select has color orange}\\
 \uttnq{add red top 4 times} :=  \uttnq{repeat 4 [add red top]}\\
 \uttnq{l white} := \uttnq{go left and add white} \\
 \uttnq{mov up 2} := \uttnq{repeat 2 [select up]}\\
 \uttnq{go up 3} := \uttnq{go up 2; go up}\\
  \hline
 \textbf{Higher level} \\\hline
 \uttnq{add red plate 6 x 7},
 \uttnq{green cube size 4},\\
 \uttnq{add green monster}, \uttnq{black 10x10x10 frame},\\
 \uttnq{flower petals}, 
 \uttnq{deer leg back},
 \uttnq{music box}, \uttnq{dancer}
\end{tabular}
  \caption{Example definitions. See CodaLab worksheet for the full leaderboard.}
\label{tab:definition_data}
\end{table}

Three modes of naturalization are outlined
in \reftab{definition_data}. For very common operations, like moving the selection,
people found \utt{select left} too verbose and shorterned this to \uttnq{l, left, >, sel l}.
One user preferred \utt{go down and right}
instead of \utt{select bot; select right}
in core and defined it as \utt{go down; go right}.
Definitions for high-level concepts tend to be whole objects
that are not parameterized (e.g., \utt{dancer}). 
The bottom plot of \reffig{learning_curves} suggests that users are
defining and using higher level concepts,
since programs become longer relative to utterances over time.

As a result of the automatic but implicit grammar induction, some concepts do
not generalize correctly. In definition head \utt{3 tall 9 wide white tower
centered here}, arguments do not match the body;
for \utt{black 10x10x10 frame}, we failed to tokenize.

\paragraph{Learned parameters.}
Training using L1 regularization, we obtained 1713 features with non-zero parameters.
One user defined many concepts consisting of a single short token,
and the Social.Author feature for that user has the most negative weight overall.
With user compatibility (Social.Friends), some pairs have large positive weights and others large negative weights.
The \utt{isolate} scoping choice (which allows easier hierarchical building) received the most positive weights,
both overall and for many users.
The 2 highest scoring induced rules correspond to \utt{add row red right 5} and \utt{select left 2}.

\paragraph{Incentives.}
Having complex structures
show that the actions in Voxelurn are expressive and that
hierarchical definitions are useful.
To incentivize this behavior,
we created a leaderboard which ranked structures
based on recency and upvotes (like Hacker News).
Over the course of 3 days, we picked three prize categories to be released daily.
The prize categories for each day were
bridge, house, animal;
 tower, monster, flower;
 ship, dancer, and castle.

To incentivize more definitions, we also track \emph{citations}.
When a rule
is used in an accepted utterance by another user,
the rule (and its author) receives a citation.
We pay bonuses to top users according to their h-index.
Most cited definitions are also displayed on the leaderboard.
Our qualitative results should be robust to the incentives scheme,
because the users do not overfit to the incentives---e.g., around 20\% of the structures
are not in the prize categories
and users define complex concepts that are rarely cited.

\section{Related work and discussion}

This work is an evolution of \citet{wang2016games},
but differs crucially in several ways:
While \citet{wang2016games} starts from scratch and relies on selecting candidates,
this work starts with a programming language (PL) and additionally relies on definitions,
allowing us to scale.
Instead of having a private language for each user,
the user community in this work shares one language.

\citet{azaria2016instructable} presents Learning by Instruction Agent (LIA),
which also advocates learning from users. They argue that developers cannot
anticipate all the actions that users want,
and that the system cannot understand the corresponding natural
language even if the desired action is built-in.
Like \citet{jia2017concepts}, \citet{azaria2016instructable}
starts with an ad-hoc set of initial slot-filling commands in natural language
as the basis of further instructions---our approach
starts with a more expressive core PL designed to interpolate with natural language.
Compared to previous work,
this work studied interactive learning in a shared community setting and
hierarchical definitions resulting in more complex concepts.


Allowing ambiguity and a flexible syntax
is a key reason
why natural language is easier to produce---this cannot be achieved by PLs such as Inform and COBOL
which look like natural language.
In this work, we use semantic parsing techniques that can handle ambiguity
\cite{zettlemoyer05ccg,zettlemoyer07relaxed,kwiatkowski10ccg,liang11dcs,pasupat2015compositional}.
In semantic parsing, the semantic representation and action space
is usually designed to accommodate the natural language that is considered constant.
In contrast, the action space is considered constant in the naturalizing PL approach,
and the language adapts to be more natural while accommodating the action space.

Our work demonstrates that interactive definitions
is a strong and usable form of supervision.
In the future, we wish to test these ideas in more domains,
naturalize a real PL, and handle paraphrasing and implicit arguments.
In the process of naturalization,
both data and the semantic grammar have important roles
in the evolution of
a language that is easier for humans to produce while still parsable by computers.

%
%
%
%
%
%
%
%
%
%

\paragraph{Acknowledgments.}
We thank our reviewers, Panupong (Ice) Pasupat for helpful suggestions and
discussions on lambda DCS,
DARPA Communicating with Computers (CwC) program under ARO
prime contract no.\ W911NF-15-1-0462,
and NSF CAREER Award no.\ IIS-1552635.

\paragraph{Reproducibility.} All code, data, and experiments for this
paper are available on the CodaLab platform: \\
{\small \url{https://worksheets.codalab.org/worksheets/0xbf8f4f5b42e54eba9921f7654b3c5c5d}}\\
and a demo: {\small \url{http://www.voxelurn.com}}

\bibliographystyle{acl_natbib}
\bibliography{refdb/all}

\end{document}